\documentclass[11pt,a4paper]{article}
\usepackage[hyperref,final]{emnlp2021}
\usepackage{times}
\usepackage{multirow}
\usepackage{amsmath}
\usepackage{booktabs}
\usepackage{graphicx}
\usepackage{pifont}
\usepackage{array}
\usepackage{enumitem}
\usepackage{subcaption}
\usepackage{bm}
\usepackage{latexsym}
\usepackage{xspace}

\usepackage{microtype}

\newif{\ifhidecomments}
\hidecommentsfalse
\ifhidecomments
    \newcommand{\rosa}[1]{}
    \newcommand{\chenhao}[1]{}
\else
    \newcommand{\chenhao}[1]{\textcolor{blue}{[#1 ---\textsc{ct}]}}
    \newcommand{\rosa}[1]{\textcolor{brown}{[#1 ---\textsc{YZ}]}}
\fi

\newcommand{\figref}[1]{Fig.~\ref{#1}}

\newcommand{\esnli}{\textsc{e-SNLI}\xspace}
\newcommand{\ehans}{\textsc{e-HANS}\xspace}
\newcommand{\hans}{\textit{HANS}\xspace}
\newcommand{\mvmt}{\textit{IND vocab, IND template}\xspace}
\newcommand{\misvmt}{\textit{OOD vocab, IND template}\xspace}
\newcommand{\mvmist}{\textit{IND vocab, OOD template}\xspace}
\newcommand{\misvmist}{\textit{OOD vocab, OOD template}\xspace}
\newcommand{\mist}{{OOD templates}\xspace}
\newcommand{\mt}{{IND templates}\xspace}
\newcommand{\misv}{{OOD vocab}\xspace}
\newcommand{\mv}{{IND vocab}\xspace}

\newcommand{\explainthenpredict}{explain-then-predict\xspace}

\title{Investigating the Effect of Natural Language Explanations on Out-of-Distribution Generalization in Few-shot NLI}

\author{Yangqiaoyu Zhou \\
  University of Chicago \\
  \texttt{zhouy1@uchicago.edu} \\\And
  Chenhao Tan \\
  University of Chicago \\
  \texttt{chenhao@uchicago.edu} \\}

\date{}

\begin{document}
\maketitle
\begin{abstract}

Although neural models have shown strong performance in datasets such as SNLI, they lack the ability to generalize out-of-distribution (OOD). In this work, we formulate a few-shot learning setup and examine the effects of natural language explanations on OOD generalization. We leverage the templates in the \hans dataset and construct templated natural language explanations for each template. 
Although generated explanations show competitive BLEU scores against groundtruth explanations, they fail to improve prediction performance.
We further show that generated explanations often hallucinate information and miss key elements that indicate the label.

\end{abstract}

\section{Introduction}
\label{sec:intro}

Thanks to recent advances in pre-trained language models~\cite{vaswani2017attention, devlin2018bert}, 
the state-of-the-art accuracy for natural language inference (NLI) can easily exceed 90\%~\citep{pilault2020conditionally}.
However, these NLI models show poor out-of-distribution (OOD) generalization. 
For instance, \citet{mccoy2019right} create a templated dataset (\hans) and find model performance to be about chance in this dataset.

While recent studies try to tackle this robustness problem from the perspectives of both the dataset and the model \citep{le2020adversarial,swayamdipta2020dataset,clark2019don},
we investigate an extra dimension of information, natural language explanations. 
Our work is motivated by the growing interest in explanations in the NLP community~\citep{camburu2018snli, rajani2019explain, alhindi2018your, stammbach2020fever}: these explanations can potentially enable models to understand the reasoning strategy beyond spurious patterns.
We focus on a few-shot learning setup because it is unrealistic to expect a large number of annotated OOD examples.

To introduce an OOD setting with natural language explanations, we construct \ehans, a dataset with natural language explanations for each template in \hans. 
By leveraging the templates in \hans, we avoid the challenges in crowdsourcing natural language explanations \citep{wiegreffe-marasovic-2021-review} and manually build an explanation dataset of high-quality.

We use an \textsc{ExplainThenPredict} framework to learn with explanations. An explanation generation model outputs an explanation for each input example, and the generated explanation is fed into a classifier along with the input example. 
While BLEU scores imply high quality of generated explanations, learning with explanations does not improve predictive performance either in-distribution or out-of-distribution. We show the generated explanations contain words in the true explanations, but they fail to reproduce important phrases and often hallucinate entities during generation.

\section{Building Natural Language Explanations for \hans}
\label{sec:datasets}

To investigate whether natural language explanations 
improve the robustness of natural language inference (NLI),
we 
build on two existing datasets: 
1) \hans, which introduces templates to generate OOD examples for robust evaluation of NLI models; 2) \esnli, which provides explanations for the Stanford Natural Language Inference (SNLI) dataset.
Our key contribution is to augment \hans by building templated natural language explanations and studying the effect of these explanations on model robustness in a few-shot learning setup.
Our dataset is available at \href{https://github.com/ChicagoHAI/hans-explanations}{https://github.com/ChicagoHAI/hans-explanations}.

\subsection{Existing Datasets}

We start by presenting details of existing datasets.

\paragraph{\hans \citep{mccoy2019right}} contains NLI examples designed to be challenging for models that 
tend to learn spurious patterns. 
It targets known heuristics for the majority of existing NLI data. For example, one heuristic assumes that a premise entails all hypotheses that are constructed using only words in the premise.
There are 3 heuristics in \hans, each containing 10 subcases. A subcase is supported by a few
templates and the dataset is constructed following these templates.

\paragraph{\esnli \citep{camburu2018snli}} 
develops free-form self-contained explanations for the true labels in SNLI using crowdsourcing.
We pretrain a model on this dataset to examine the effect of pretraining.
There are three explanations collected for each example in the validation dataset, and we use the first explanation.
We do not use the test set of \esnli.

\begin{table}
    \small
    \centering
    \begin{tabular}{p{0.45\textwidth}}
    \toprule
    \textbf{Premise}: the psychologist by the programmers saw the essayist. \\
    \textbf{Hypothesis}: the psychologist saw the essayist. \\
    \textbf{Explanation}: the psychologist by the programmers is still the psychologist. \\
    \bottomrule
\end{tabular}
\caption{An example from \ehans. 
Average length of premise and hypothesis are 8.8 and 4.4 tokens. 
Average length of natural language explanation is 13.3 tokens.
}
\label{tab:examples}
\end{table}

\subsection{Templated Natural Language Explanations for \hans} 

We build natural language explanations for \hans to examine whether explanations can help
models when facing this challenging corpus.  
As \hans is constructed with templates, we develop templates for natural language explanations accordingly.
They explain the reasons for the true label in human language.

Table~\ref{tab:examples} shows an example of the proposed explanations (more examples are in Appendix \ref{sec:appendix_ehans}). 
In addition to developing these templated explanations, 
we expand the original HANS vocabulary in terms of its nouns, verbs, adjectives, and adverbs to increase difficulty.
This allows us to examine the effect of unseen words.

\section{Experiments}
\label{sec:setup}

\subsection{Few-Shot Learning Set-up} 

To investigate whether natural language explanations improve the robustness of NLI models, 
we look at a few-shot learning setting.
We focus on this setting since in practice one may have little or no access to OOD instances.
We are interested in the following questions:
\begin{enumerate}[leftmargin=*,itemsep=-4pt,topsep=0pt]
\item whether the model trained from in-distribution examples can generalize 
to unseen templates and words,
\item how many samples are enough for learning,
\item whether pretraining on \esnli improves generalization on \hans,
\item and most importantly, what is the effect of explanations.
\end{enumerate}
We use 5-fold cross validation by splitting 118 templates randomly into 5 folds.
We generate $k$ samples for each training template using \ehans explanation templates. 
We then build a corresponding development set that contains $0.2k$ samples of 
each 
training template,
so that the development set is 20\% the size of the training set and does not include any unseen template.
This setup ensures that the size of the development set is realistic \citep{kann2019towards}.

Finally, we build test instances in the following categories to evaluate the performance of the models both in-distribution and out-of-distribution:
\begin{itemize}[itemsep=-4pt,leftmargin=*,topsep=2pt]
    \item \mvmt. Both the templates and the vocabulary are matched with the training set. We expect the performance to grow steadily as $k$ increases.
    \item \misvmt. We use the same templates as the training set, but use unseen words to generate this test set. The challenge lies in understanding unseen words.
    \item \mvmist. We use the unseen templates and the same vocabulary as the training set. The challenge lies in understand the logic encoded in unseen templates.
    \item \misvmist. Finally, we generate the test data with both the unseen templates and the unseen words. 
\end{itemize}
We use the same test sets (300 examples for each template) 
to examine how the models' performance changes as $k$ increases.

\subsection{Models} 

\begin{figure*}
    \centering
    \begin{subfigure}[t]{0.23\textwidth}
        \centering
        \includegraphics[width=\textwidth]{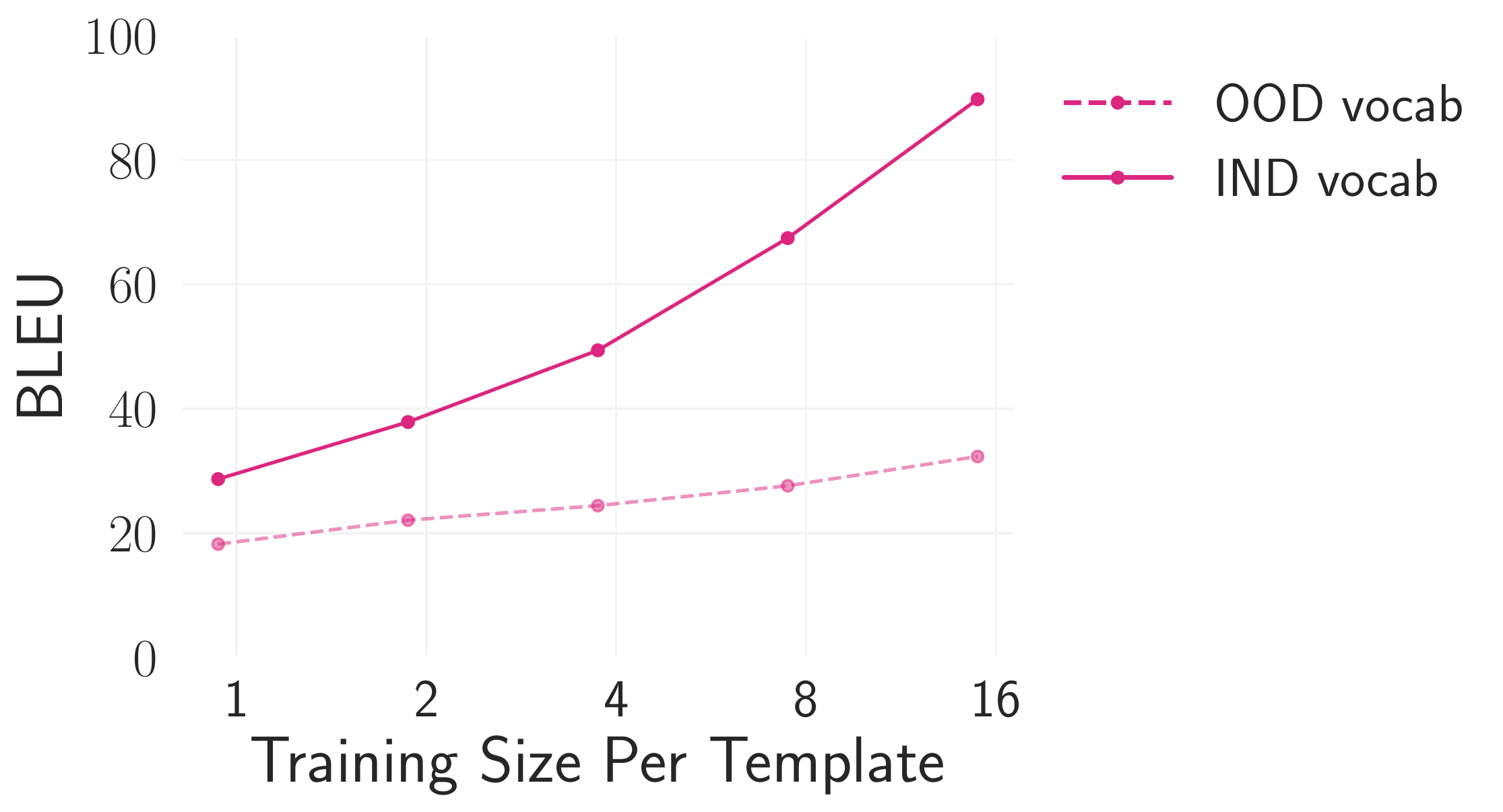}
        \caption[BERT matched temp]%
        {{\mt with BERT}}
        \label{fig:abundant_temp_bleu_bert_mt}
    \end{subfigure}
    \hfill
    \begin{subfigure}[t]{0.23\textwidth}
        \centering
        \includegraphics[width=\textwidth]{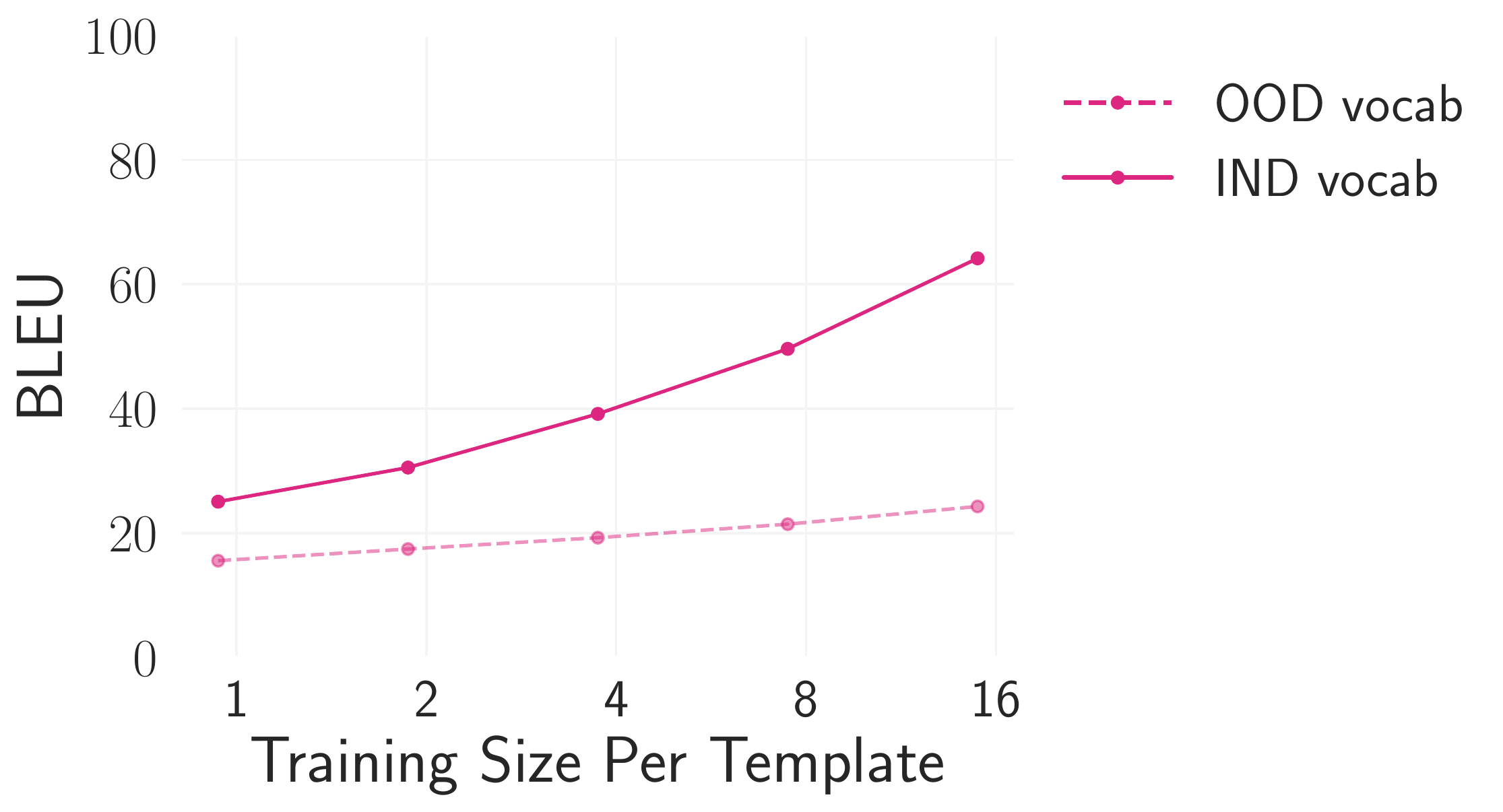}
        \caption[BERT mismatched temp]%
        {{\mist with BERT}}
        \label{fig:abundant_temp_bleu_bert_mist}
    \end{subfigure}
    \hfill
    \begin{subfigure}[t]{0.23\textwidth}
        \centering
        \includegraphics[width=\textwidth]{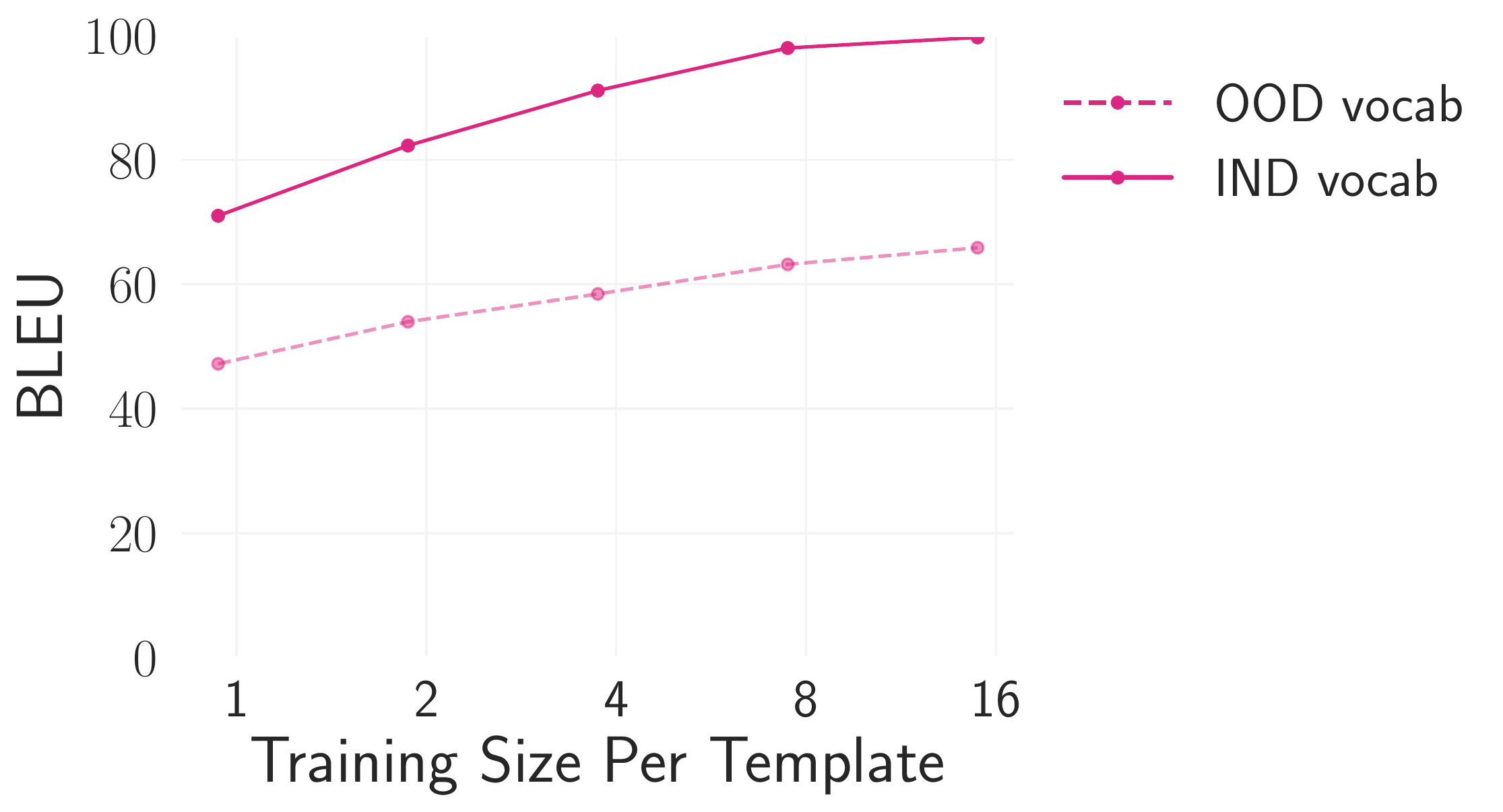}
        \caption[ESNLI matched temp]%
        {{\mt with ESNLI}}
        \label{fig:abundant_temp_bleu_esnli_mt}
    \end{subfigure}
    \hfill
    \begin{subfigure}[t]{0.23\textwidth}
        \centering
        \includegraphics[width=\textwidth]{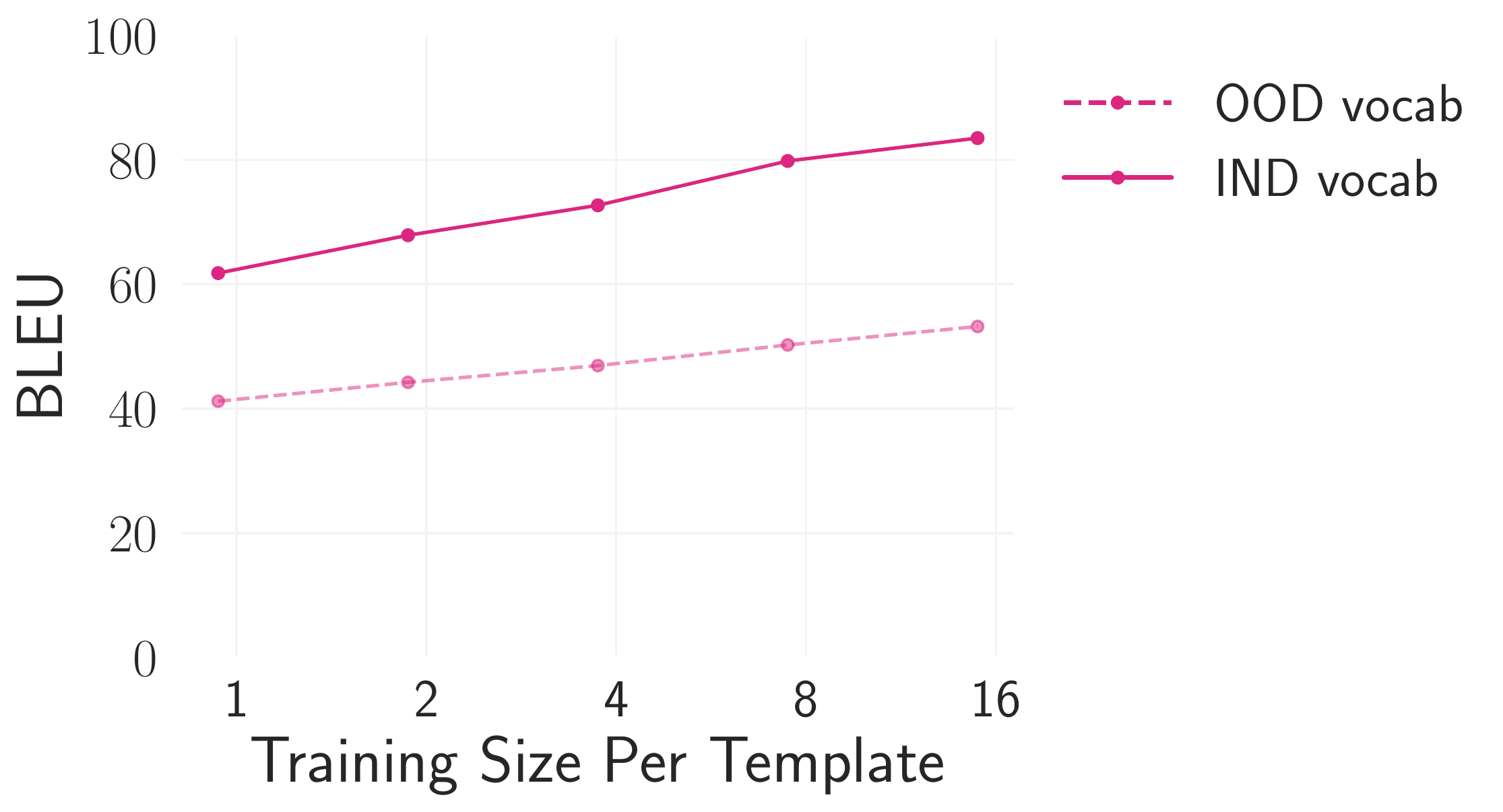}
        \caption[ESNLI mismatched temp]%
        {{\mist with ESNLI}}
        \label{fig:abundant_temp_bleu_esnli_mist}
    \end{subfigure}
    \caption[EffectOfExplAndDataSize]%
    {%
    $x-$axis shows the number of samples per template, while $y-$axis shows the BLEU score. 
      BLEU scores are high for \mvmt instances. Although BLEU drops substantially for both BERT and the \esnli pretrained model under \misv and \mist, it is still decent (above 40 with \esnli).
    }
    \label{fig:abundant_temp_bleu}
\end{figure*}

We adapt the \textsc{ExplainThenPredict} architecture introduced by 
\citet{camburu2018snli} in our experiments. It consists of \textbf{a generation model} and \textbf{a classification model}.
The generation model produces an explanation given an input premise and hypothesis pair. 
This generated explanation and the original input are fed into the classifier 
for label prediction.
Our framework slightly differs from \citet{camburu2018snli} 
in that their classifier 
only takes explanation as input for the classifier.

The explanation generation model follows an encoder-decoder framework.
Both the encoder and the decoder use the BERT model, but the decoder uses a masking mechanism so that 
it predicts the next word considering only all the preceding words in both training and testing phases. 
Our generation model obtains close to SoTA performance on e-SNLI, comparing against WT5 (33.15 vs. 33.7 in BLEU) \citep{narang2020wt5}.

The classification model is a BERT sequence classifier, where a linear layer is applied to the pooled output of BERT encodings (i.e., embedding of the CLS token). 

\subsection{Experimental Setup}

We used $k = 1, 2, 4, 8, 16$ to generate the training data. 
The explanation generator trains on groundtruth explanations. 
For all of our models, we saved the model with the best validation 
performance during training and did not tune other hyperparameters. 
We used a training batch size 16 and a learning rate 5e-5 for the explanation generator,
and we used a training batch size 128 and learning rate 2e-5 for the classifier. 

\paragraph{Model comparisons.} To test if explanations help with learning, 
we compare with a baseline that only includes the classifier component with the input premise, hypothesis pair (hence ``{\em label-only}'').
We also consider a baseline that does not update with the $k$ samples in the training set (hence ``{\em no training}'') and a majority baseline (``{\em majority}''). 

In addition, we compare the vanilla BERT model
with a BERT model fine-tuned on \esnli during both generation and classification to investigate whether pretraining on \esnli helps with the \hans task. 

\section{Results}
\label{sec:results}

We first look at the quality of generated explanations using BLEU. 
Although the generated explanations 
match groundtruth explanations well based on BLEU,
they affect downstream classification negatively in our few-shot learning set-up.
We further examine the generated explanations to understand why the predictive performance drops when adding natural language explanations.

\subsection{Quality of Explanations based on BLEU}

\begin{figure*}
    \centering
    \begin{subfigure}[t]{0.23\textwidth}
        \centering
        \includegraphics[width=\textwidth]{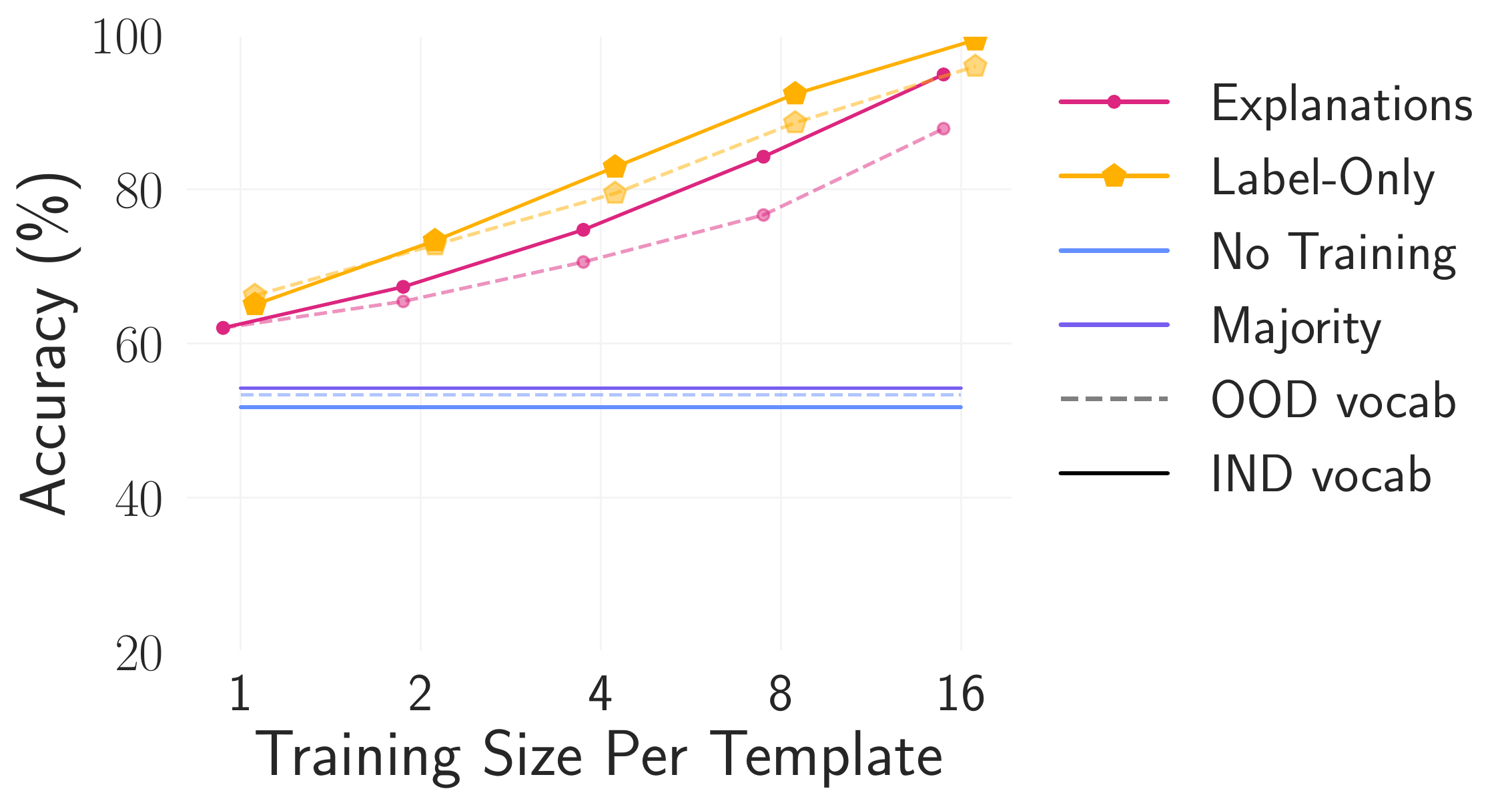}
        \caption[BERT matched temp]%
        {{\mt with BERT}}
        \label{fig:abundant_temp_acc_bert_mt}
    \end{subfigure}
    \hfill
    \begin{subfigure}[t]{0.23\textwidth}
        \centering
        \includegraphics[width=\textwidth]{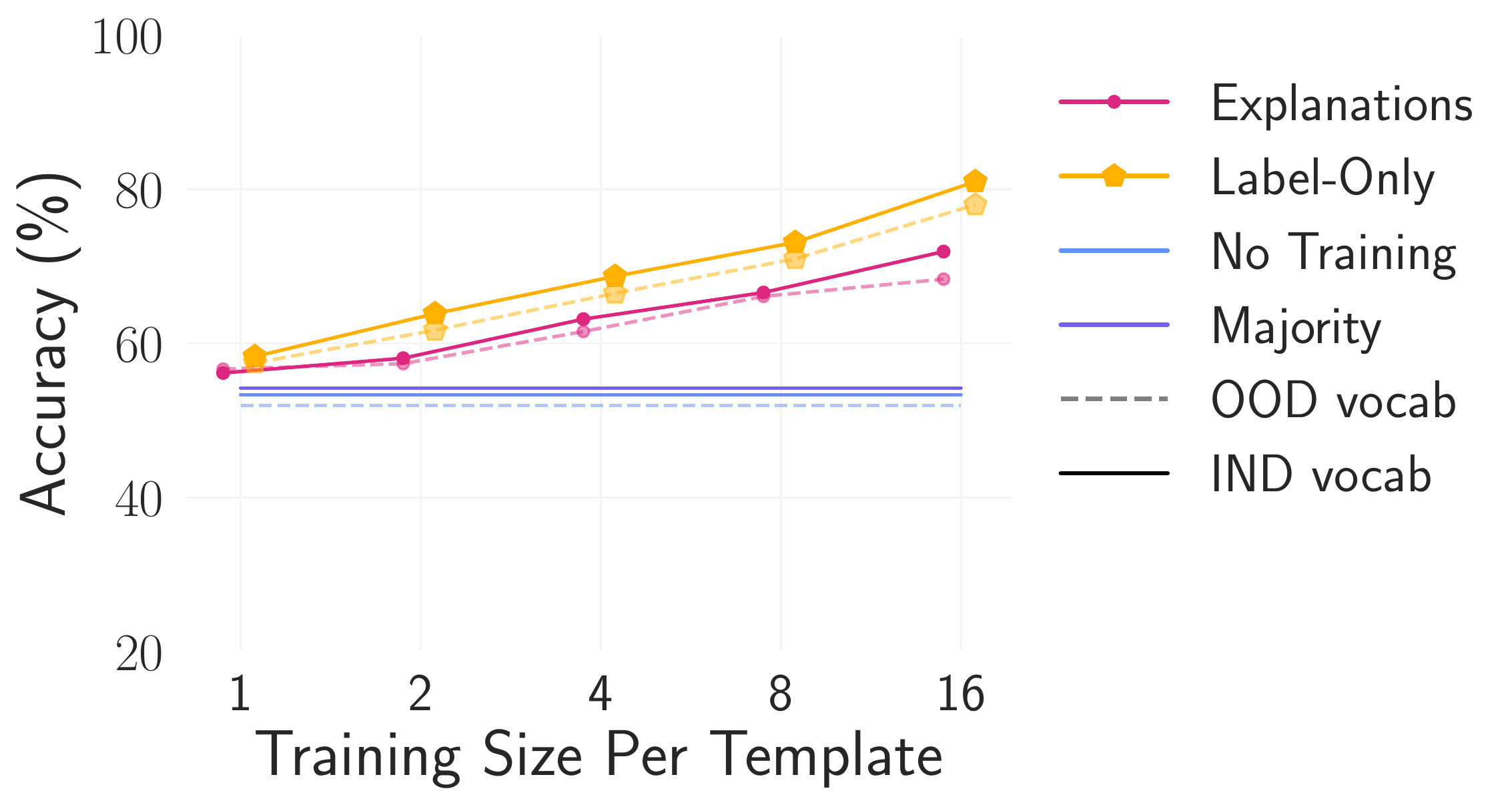}
        \caption[BERT mismatched temp]%
        {{\mist with BERT}}
        \label{fig:abundant_temp_acc_bert_mist}
    \end{subfigure}
    \hfill
    \begin{subfigure}[t]{0.23\textwidth}
        \centering
        \includegraphics[width=\textwidth]{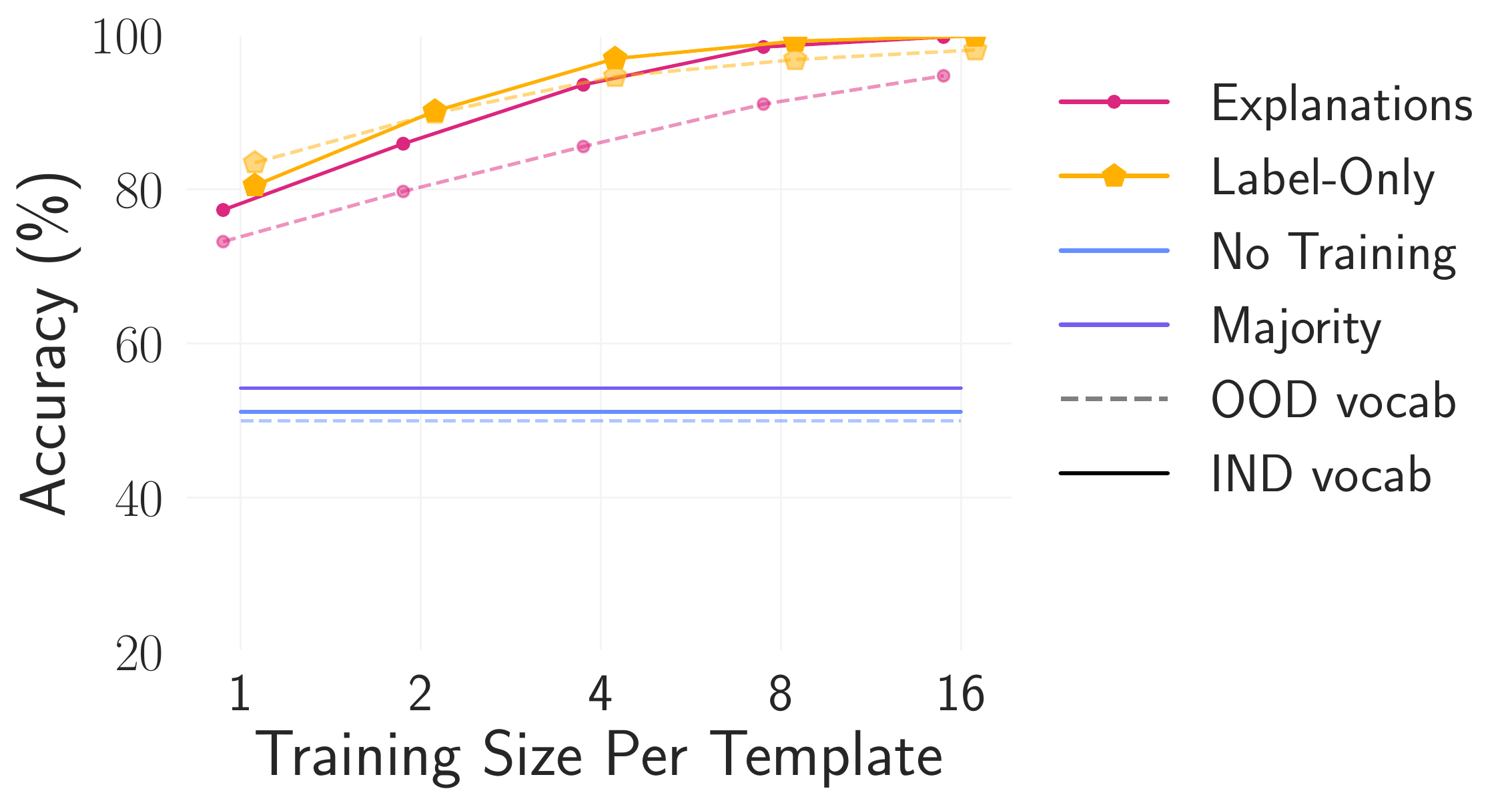}
        \caption[ESNLI matched temp]%
        {{\mt with ESNLI}}
        \label{fig:abundant_temp_acc_esnli_mt}
    \end{subfigure}
    \hfill
    \begin{subfigure}[t]{0.23\textwidth}
        \centering
        \includegraphics[width=\textwidth]{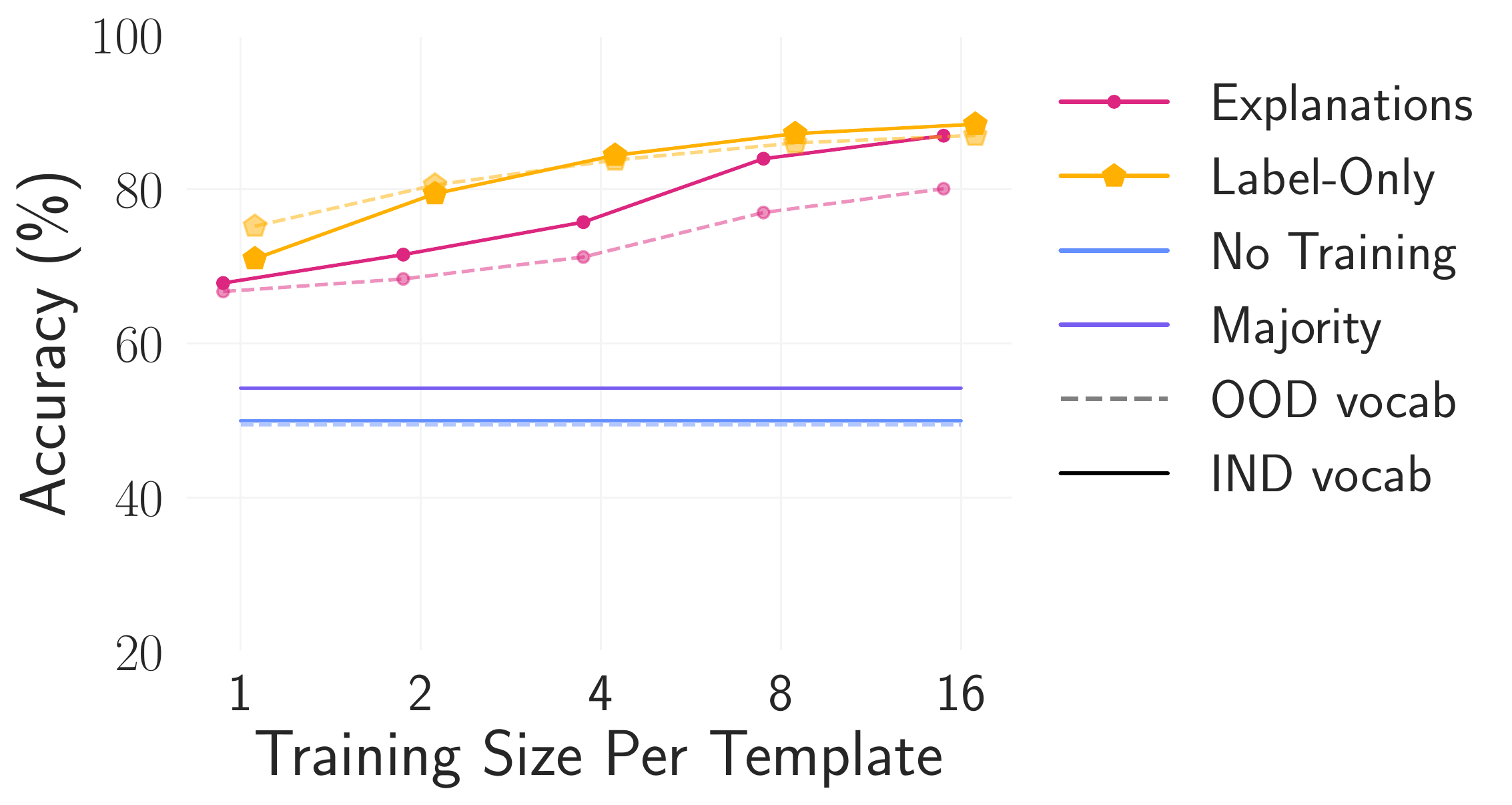}
        \caption[ESNLI mismatched temp]%
        {{\mist with ESNLI}}
        \label{fig:abundant_temp_acc_esnli_mist}
    \end{subfigure}
    \caption[EffectOfExplAndDataSize]%
    {$x$-axis shows the number of samples per template, while $y-$axis shows the accuracy in label prediction. 
    Learning from explanations is always below the label-only baseline.
    }
    \label{fig:abundant_temp_acc}
\end{figure*}

Generated explanations achieve high BLEU scores based on our groundtruth templated explanations (Figure \ref{fig:abundant_temp_bleu}). \mv explanations can achieve BLEU scores greater than 90 on \mt and 60 on \mist
when $k=16$. 
Even \misv explanations can achieve BLEU scores close to 20.
In general, the performance grows steadily as $k$ increases for both IND and OOD.

While the BLEU scores can be quite high, OOD generalization remains a challenge.
Unseen vocabulary and templates (\figref{fig:abundant_temp_bleu_bert_mist}, \figref{fig:abundant_temp_bleu_esnli_mist}) increase the difficulty in explanation generation.  
That said, pretraining on \esnli improves generation quality for both IND and OOD cases. This improvement on OOD generalization is likely due to exposure to other data during pretraining.

We use BLEU to evaluate the quality of generated explanations with regard to groundtruth explanations because it is a commonly used metric to evaluate natural language explanations \citep{camburu2018snli,rajani2019explain}. 

\subsection{Predictive Performance}

Despite the high BLEU scores, learning with the generated explanations does not help the classification task (\figref{fig:abundant_temp_acc}). 
Learning with explanations consistently performs worse than the label-only baseline under both IND and OOD testing scenarios. Pretraining on \esnli does not change this observation either. 

The only positive result we find is that pretraining helps with OOD generalization.
Models pretrained on \esnli give better results than plain BERT (\figref{fig:abundant_temp_acc}). This finding aligns with the positive effect of pretraining on explanation generation. 

We also observe that testing on groundtruth explanations boosts performance drastically. 
This suggests that groundtruth explanations give clues for the label, but generated explanations do not capture this information.

\subsection{Why Explanations Do Not Help?}

\begin{figure*}[t]
    \centering
    \begin{subfigure}[t]{0.23\textwidth}
    \includegraphics[width=\textwidth]{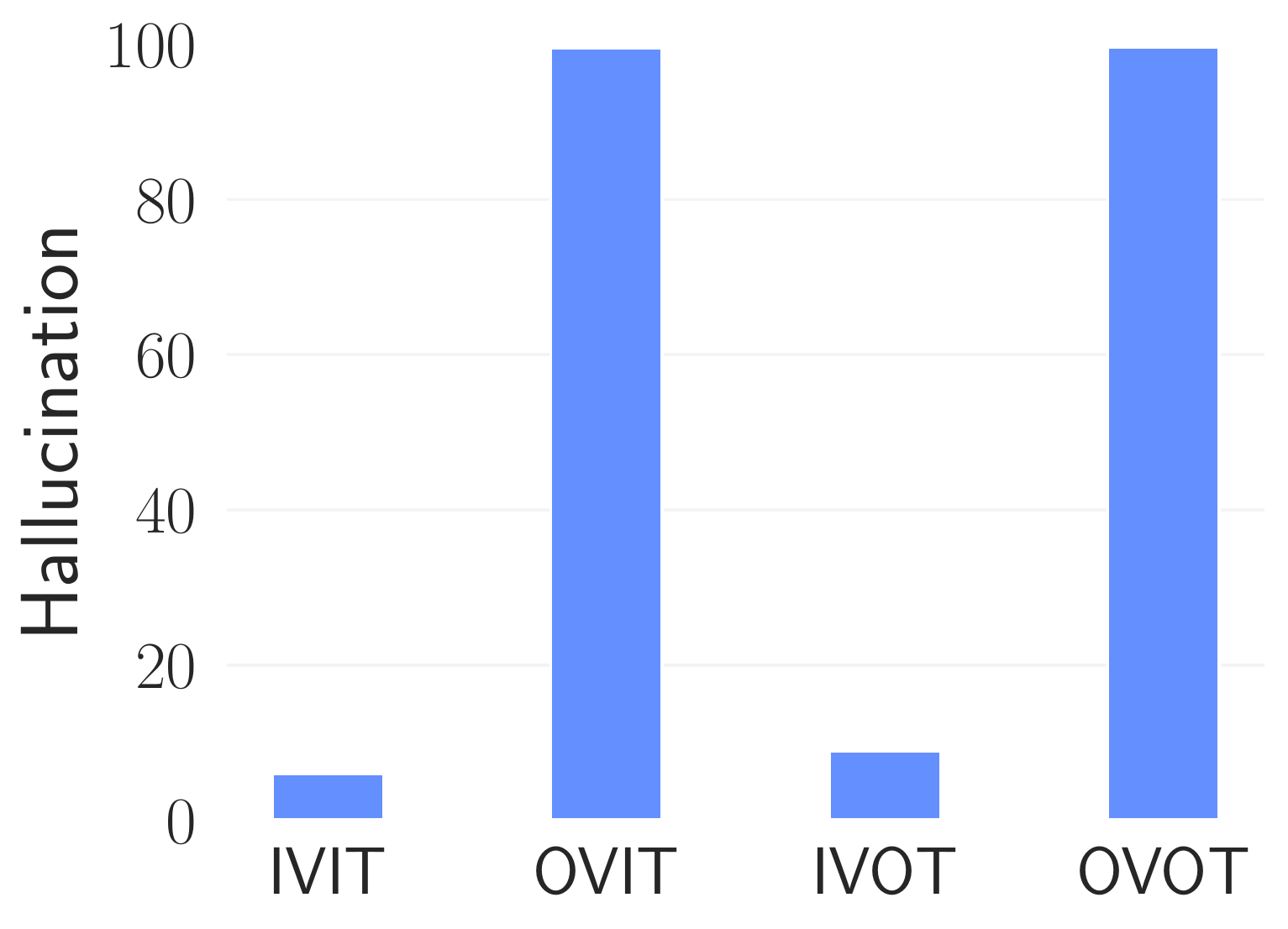}
    \caption{\% of explanations with hallucinated entities. (BERT)}
    \label{fig:k=16_hallucination_bar_chart_bert}
    \end{subfigure}
    \hfill
    \begin{subfigure}[t]{0.23\textwidth}
    \includegraphics[width=\textwidth]{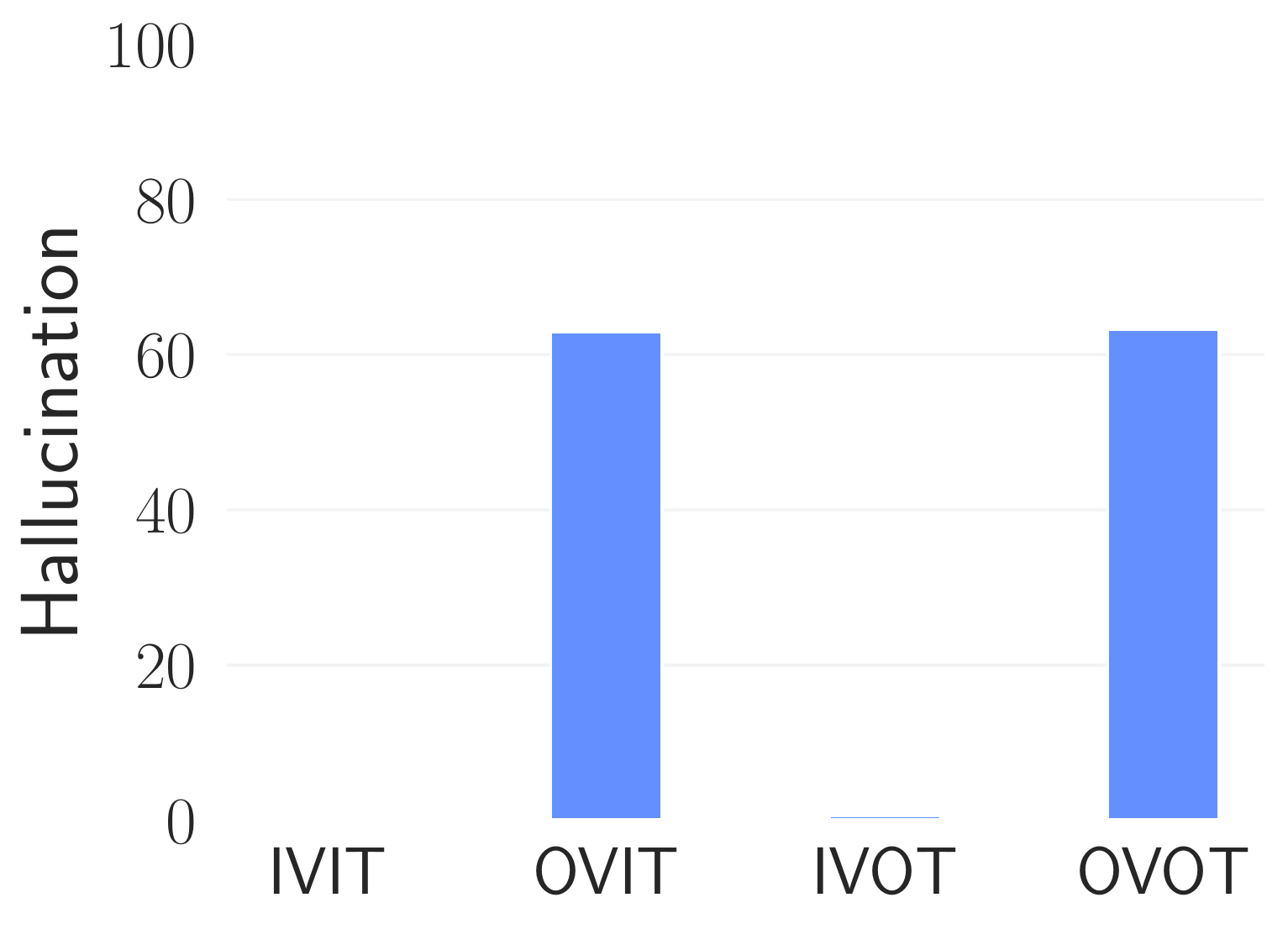}
    \caption{\% of explanations with hallucinated entities. (\esnli)}
    \label{fig:k=16_hallucination_bar_chart_esnli}
    \end{subfigure}
    \hfill
    \begin{subfigure}[t]{0.23\textwidth}
    \includegraphics[width=\textwidth]{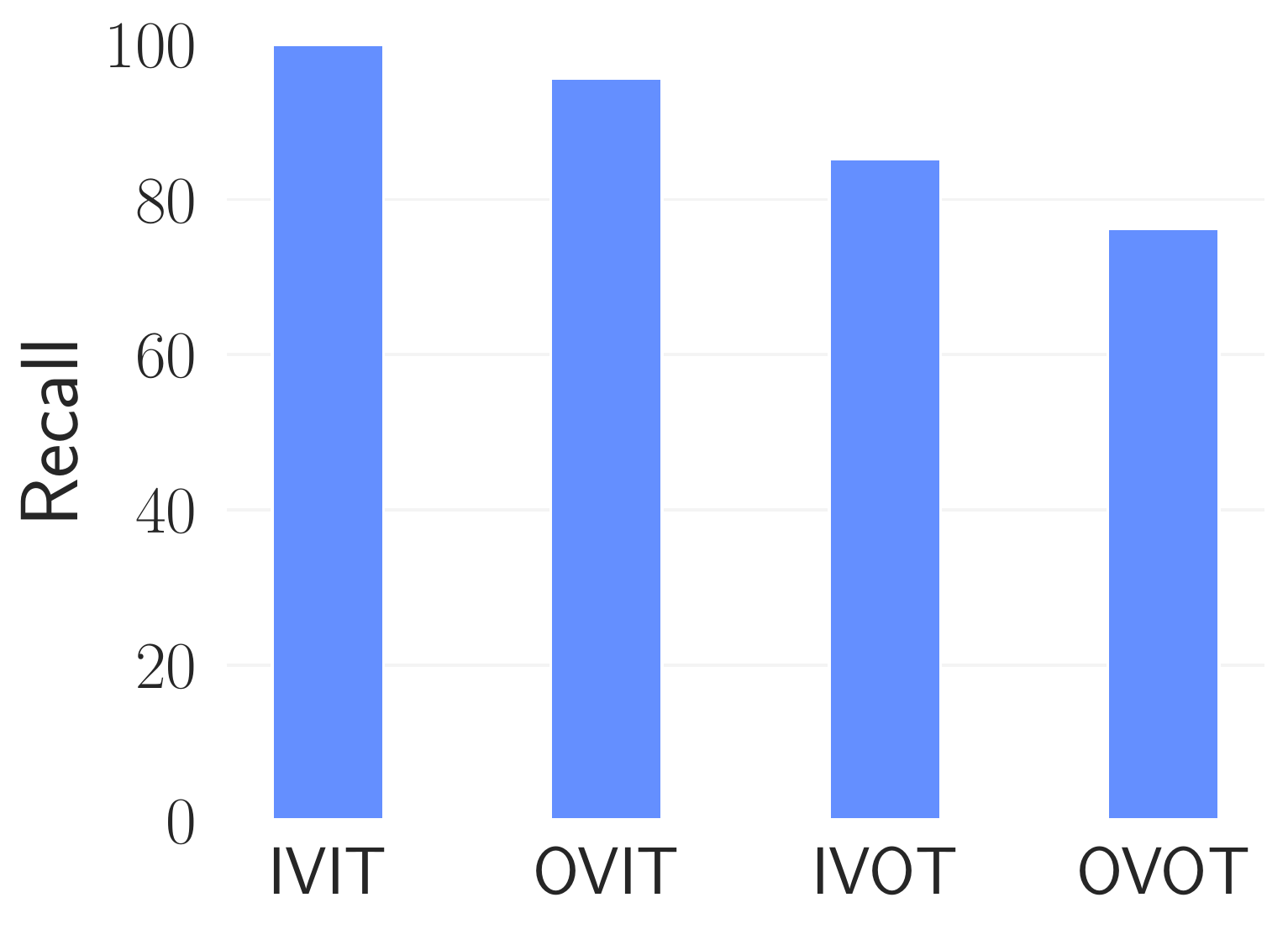}
    \caption{Recall of generating ``we do not know''. (BERT)}
    \label{fig:k=16_wdnk_bar_chart_bert}
    \end{subfigure}
    \hfill
    \begin{subfigure}[t]{0.23\textwidth}
    \includegraphics[width=\textwidth]{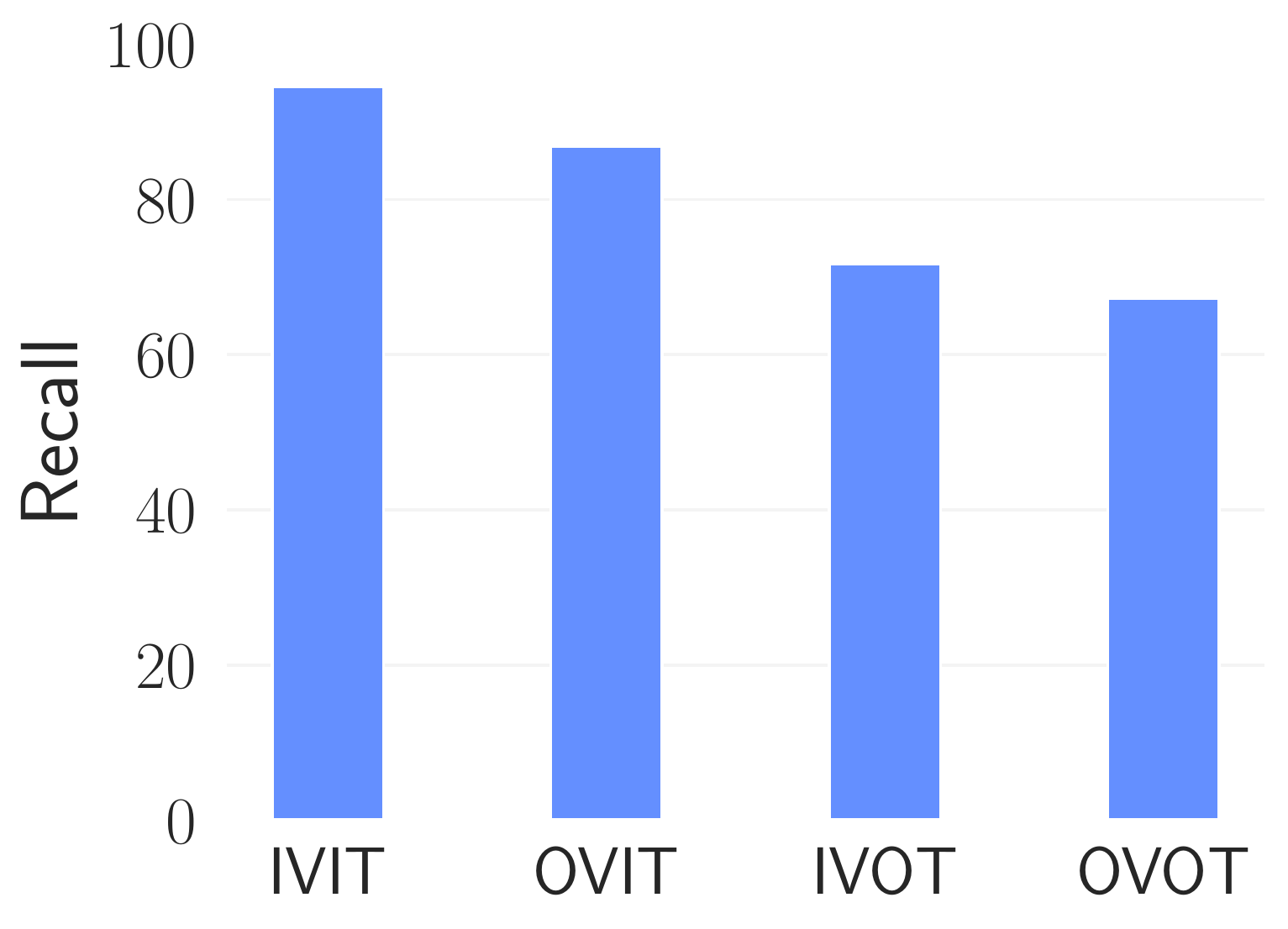}
    \caption{Recall of generating ``we do not know''. (\esnli)}
    \label{fig:k=16_wdnk_bar_chart_esnli}
    \end{subfigure}
    \caption{\figref{fig:k=16_hallucination_bar_chart_bert} and \figref{fig:k=16_hallucination_bar_chart_esnli} show that the BERT model and the \esnli-pretrained model (trained with $k=16$) hallucinate for \misv.
    \figref{fig:k=16_wdnk_bar_chart_bert} and \figref{fig:k=16_wdnk_bar_chart_esnli} suggest that the explanations fail to include ``we do not know'' for instances with the non-entailment label for \misv and \mist (with $k=16$).
    }
    \label{fig:k=16_expl_quantitative}
\end{figure*}

\begin{figure*}[t]
    \centering
    \begin{subfigure}[t]{0.23\textwidth}
    \includegraphics[width=\textwidth]{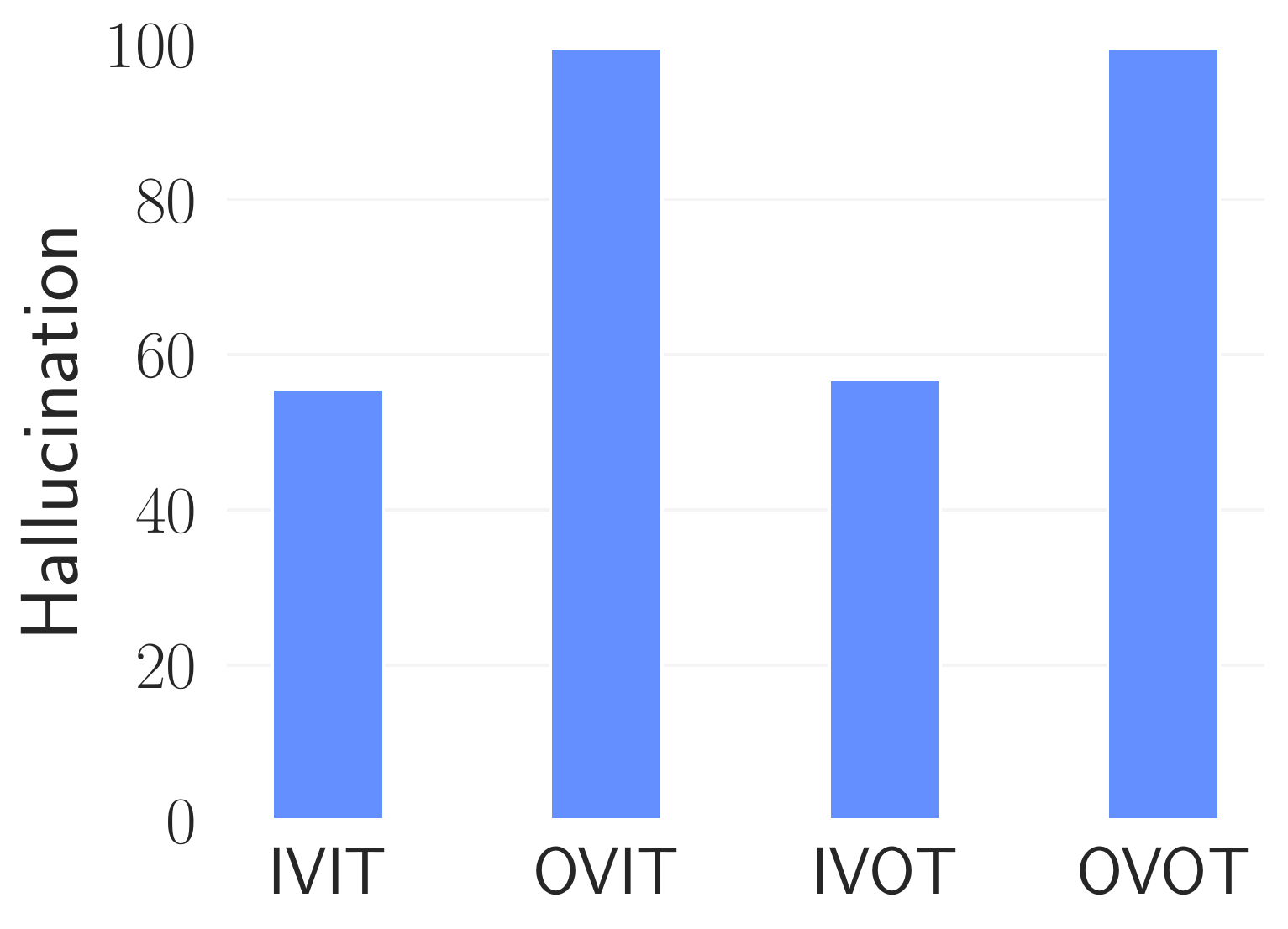}
    \caption{\% of explanations with hallucinated entities. (BERT)}
    \label{fig:k=4_hallucination_bar_chart_bert}
    \end{subfigure}
    \hfill
    \begin{subfigure}[t]{0.23\textwidth}
    \includegraphics[width=\textwidth]{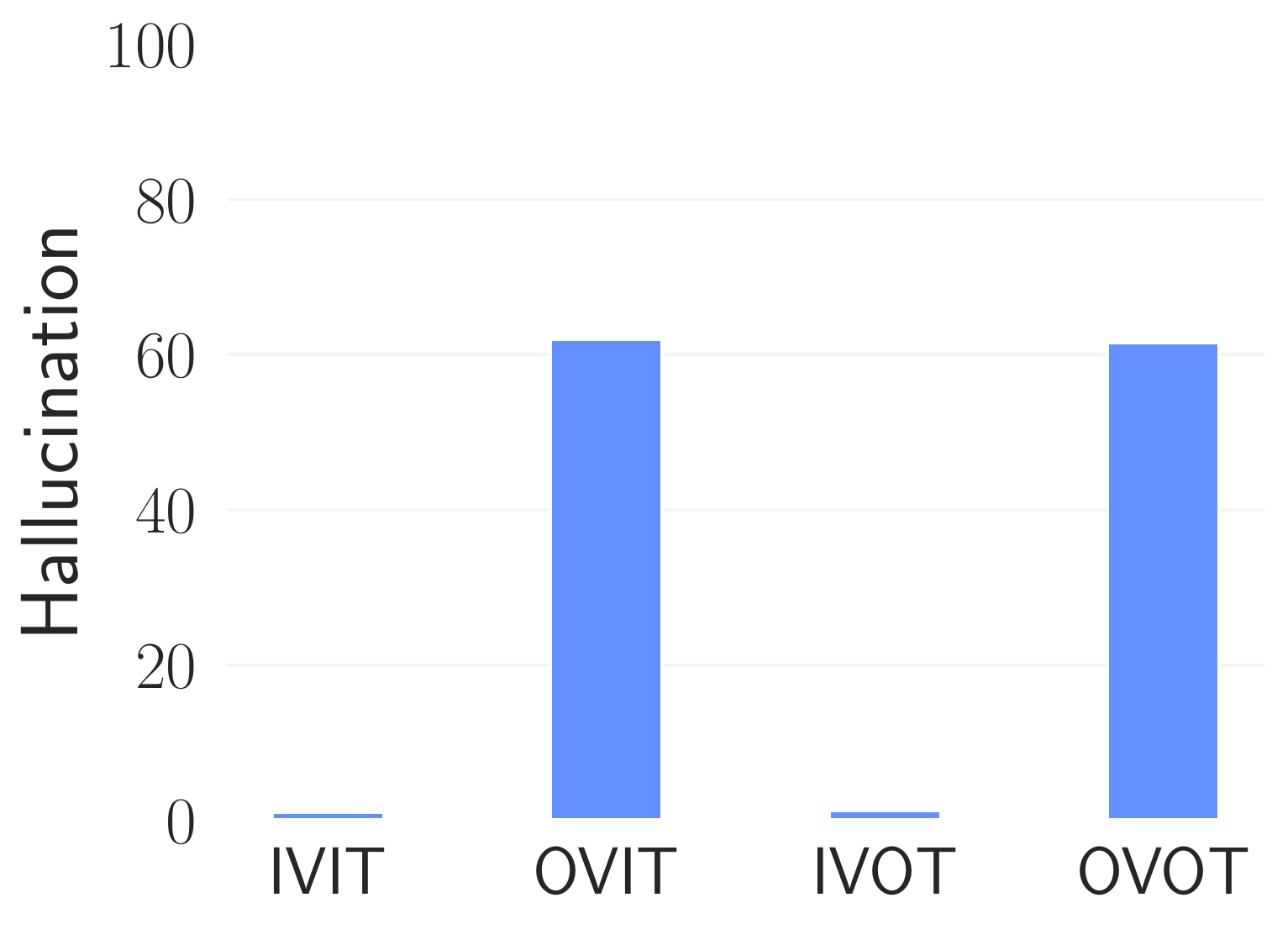}
    \caption{\%  of explanations with hallucinated entities. (\esnli)}
    \label{fig:k=4_hallucination_bar_chart_esnli}
    \end{subfigure}
    \hfill
    \begin{subfigure}[t]{0.23\textwidth}
    \includegraphics[width=\textwidth]{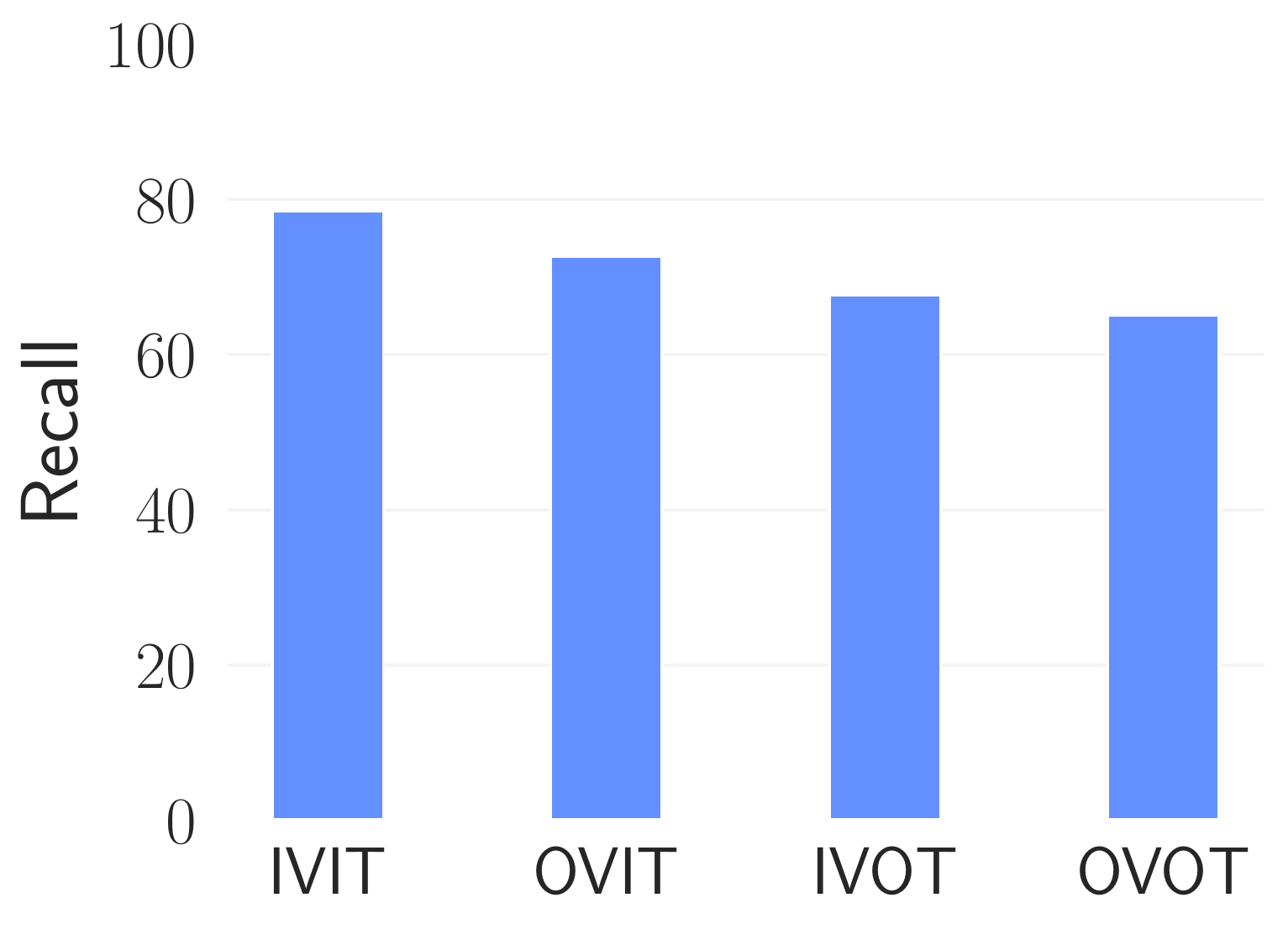}
    \caption{Recall of generating ``we do not know''. (BERT)}
    \label{fig:k=4_wdnk_bar_chart_bert}
    \end{subfigure}
    \hfill
    \begin{subfigure}[t]{0.23\textwidth}
    \includegraphics[width=\textwidth]{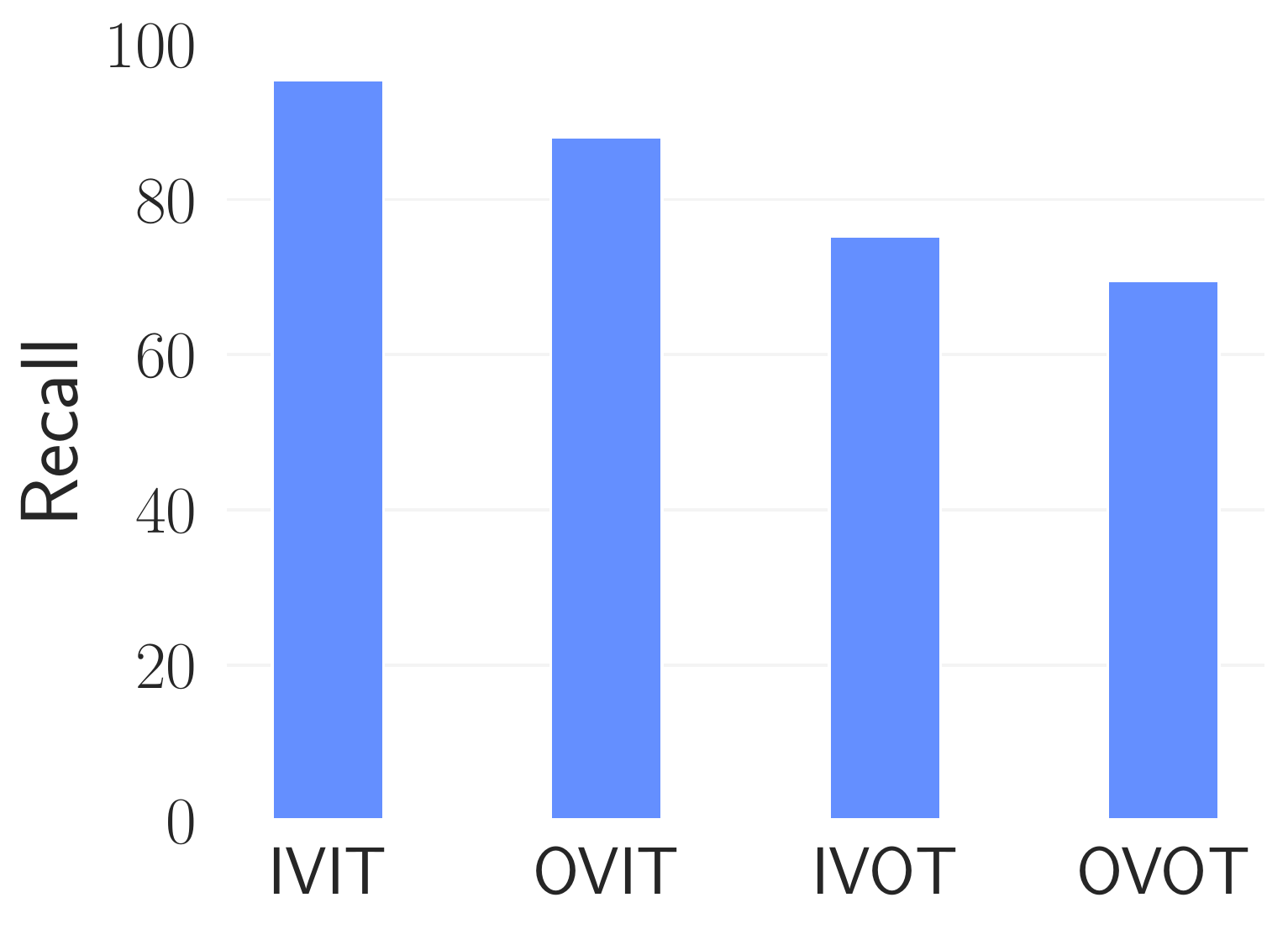}
    \caption{Recall of generating ``we do not know''. (\esnli)}
    \label{fig:k=4_wdnk_bar_chart_esnli}
    \end{subfigure}
    \caption{\figref{fig:k=4_hallucination_bar_chart_bert} and \figref{fig:k=4_hallucination_bar_chart_esnli} show that both the BERT model and \esnli-pretrained model (trained with $k=4$) hallucinate for \misv, and the hallucination rate is slightly worse for \mist.
    Similarly, \figref{fig:k=4_wdnk_bar_chart_bert} and \figref{fig:k=4_wdnk_bar_chart_esnli} suggest that the explanations fail to include ``we do not know'' for instances with the non-entailment label for \misv and \mist (with $k=4$).
    }
    \label{fig:k=4_expl_quantitative}
\end{figure*}

To understand why explanations are not helpful, 
we introduce two new metrics to evaluate the effectiveness of explanation generation.
We measure how often the generated explanations contain \textbf{hallucinated entities}, professions (i.e., people) and locations that do not show up in the input, and we measure how well the good label indicator word \textbf{``we do not know''} is generated. We present results on explanations generated by the BERT model 
and the \esnli-pretrained model.

\begin{table}[t]
\small
\centering
\begin{tabular}{p{0.45\textwidth}}
\toprule
\mvmt ($k=4$) \\
\textbf{Premise}: the managers who the baker addressed brought the technician. \\
\textbf{Hypothesis}: the baker addressed the managers. \\
\textbf{Original explanations}: who in who the baker addressed refers to the managers. \\
\textbf{BERT explanations}: the artisans that addressed the baker are still the managers. \\
\bottomrule
\mvmt ($k=16$) \\
\textbf{Premise}: the analysts in front of the programmers affected the scientist. \\
\textbf{Hypothesis}: the analysts affected the scientist. \\
\textbf{Original explanations}: the analysts in front of the programmers are still the analysts. \\
\textbf{BERT explanations}: the analysts in front of the programmers are still the analysts. \\
\bottomrule
\misvmt ($k=16$) \\
\textbf{Premise}: the chaplains near the singer needed the author. \\
\textbf{Hypothesis}: the chaplains needed the author. \\
\textbf{Original explanations}: the chaplains near the singer are still the chaplains. \\
\textbf{BERT explanations}: the psychologists are in front of the musician and the strategists helped the writer, we do not know whether the illustrators helped the writer. \\
\bottomrule
\end{tabular}
\caption{Example generated explanations for \mt cases by the BERT model trained with $k=4, 16$. More examples are in Appendix \ref{sec:appendix_example_generation}.}
\label{tab:generated_expl}
\end{table}

An explanation contains a hallucinated entity if there is an entity that never show up in the original input. These hallucinated entities will likely hinder predictive performance when models make predictions based on generated explanations. 
We only count hallucinated professions and locations to avoid false positives due to synonyms used in explanations. That is, we use a conservative estimate on hallucinated keywords in generated explanations by only counting people and locations.
We find that hallucinated entities are 
almost always generated in \misv cases by the BERT model (99\% of explanations consist of entities that do not exist in the premise and the hypothesis) and the hallucination rate is also high (around 60\%) for the \esnli-pretrained model. However, the hallucination rate is much lower for \mv cases (\figref{fig:k=16_expl_quantitative}, \figref{fig:k=4_expl_quantitative}): it is close to 0 when $k=16$ and for \esnli-pretrained model. But when $k=4$, we observe a high hallucination rate ($>50$\%) for \mv cases (\figref{fig:k=4_hallucination_bar_chart_bert}). 
We also notice that pretraining on \esnli leads to models with much lower hallucination rates for all test cases.

``We do not know'' is a predictive phrase because it is only present in non-entailment examples.
We find that when generated explanations contain ``we do not know'', so do the corresponding groundtruth explanations (in other words, precision is 100\%).
However, when ``we do not know'' is in the groundtruth explanations, it is not necessarily always generated, so the recall is not perfect (\figref{fig:k=16_wdnk_bar_chart_bert}). In fact, recall decreases as we switch to harder test cases. \mist also has greater negative impact than \misv on recall.
Finally, we look closely at some of the generated explanations (Table \ref{tab:generated_expl}). 
We observe that models struggle to learn the templates even for the \mt case. In the easiest case (\mvmt), although the explanation uses the right template when $k=16$, it uses a wrong template when $k=4$. Once we switch from \mvmt to \misvmt, even the $k=16$ models fail to learn which template should be used to generate explanations.

\section{Conclusion}
\label{sec:conclusion}

We construct a HANS-based dataset with explanations. On this dataset, we find natural language explanations do not help few-shot NLI to generate out-of-domain under an \textsc{ExplainThenPredict} framework. While the genearted explanations obtain high BLEU scores, they do not learn information crucial for downstream classification. Our generation model is close to the SoTA model, yet it still generates nonsensical explanations. Better metrics for explanation evaluation and explanation generation models are key to success for learning with natural language explanations to be effective.

\section*{Acknowledgments} 
We thank anonymous reviewers for their valuable feedbacks. We thank members of the Chicago Human+AI Lab for their insightful suggestions. We thank Tom Mccoy, one author for the HANS paper, for a detailed explanation on their data when we reached out. We thank techstaff members at the University of Chicago CS department for their technical support. 
This work is supported in part by research awards from Amazon, IBM, Salesforce, and NSF IIS-2126602.

\bibliographystyle{acl_natbib}
\bibliography{refs}

\newpage
\appendix

\section{Replicability Details}

We pretrain the BERT model on \esnli for 5 epochs and evaluate at every epoch. The model with best dev performance is saved as the final \esnli model that we use as the initial model in few-shot learning.

On the \ehans dataset, we run 2000 steps to train the generation model and evaluate every 200 steps. We choose this number because the best dev performance is usually achieved within 2000 steps.
As for the \explainthenpredict classifier, we run 200 training steps and evaluate every 4 steps because the model quickly reaches best dev performance as training starts.
On the other hand, label-only classifier takes more steps in learning. We train for 1000 steps and evaluate every 50 steps.

It takes around 30 minutes to train a generation model and 10 minutes to train a classification model on our machine (with 4 gpus).

\newpage

\section{Samples from \ehans}
\label{sec:appendix_ehans}

\begin{table}[ht]
    \small
    \centering
    \begin{tabular}{p{0.45\textwidth}}
    \toprule
    \textbf{Premise}: the scientists that talked thanked the psychotherapists. \\
    \textbf{Hypothesis}: the scientists thanked the psychotherapists. \\
    \textbf{Explanation}: the scientists that talked are still the scientists. \\
    \midrule
    \textbf{Premise}: if the psychologists ran, the programmers existed. \\
    \textbf{Hypothesis}: the psychologists ran. \\
    \textbf{Explanation}: the programmers existed if the psychologists ran, we do not know whether the psychologists ran. \\
    \midrule
    \textbf{Premise}: though the president voted, the musician existed. \\
    \textbf{Hypothesis}: the president voted. \\
    \textbf{Explanation}: though suggests the president voted happened.\\
    \midrule
    \textbf{Premise}: the scientist was addressed by the psychotherapist. \\
    \textbf{Hypothesis}: the psychotherapist addressed the scientist. \\
    \textbf{Explanation}: addressed is the active form of was addressed by, so we swap the scientist and the psychotherapist. \\
    \midrule
    \textbf{Premise}: if the directors addressed the illustrators, the president listened. \\
    \textbf{Hypothesis}: the directors addressed the illustrators. \\
    \textbf{Explanation}: the president listened if the directors addressed the illustrators, we do not know whether the directors addressed the illustrators. \\
    \bottomrule
\end{tabular}
\caption{Examples from \ehans. 
}
\label{tab:appendix_examples}
\end{table}

\newpage

\section{Example Generated Explanations}
\label{sec:appendix_example_generation}

Examples of generated explanations on different test types.

\begin{table}[h]
\small
\centering
\begin{tabular}{p{0.45\textwidth}}
\toprule
\mvmt \\
\textbf{Premise}: the musician by the psychiatrists offended the strategists. \\
\textbf{Hypothesis}: the musician offended the strategists. \\
\textbf{Original explanations}: the musician by the psychiatrists is still the musician. \\
\textbf{BERT explanations}: the strategists offended the musician and the strategists are by the musician, we do not know whether the strategists offended the musician. \\
\midrule
\misvmt \\
\textbf{Premise}: the administrators near the penciller supported the lyricist. \\
\textbf{Hypothesis}: the administrators supported the lyricist. \\
\textbf{Original explanations}: the administrators near the penciller are still the administrators. \\
\textbf{BERT explanations}: the musician is the subject of the relative clause that addressed the illustrators and the programmers encouraged the programmer, we do not know whether the planners encouraged the illustrators. \\
\midrule
\mvmist \\
\textbf{Premise}: the scientists who affected the colorists helped the psychotherapists. \\
\textbf{Hypothesis}: the scientists helped the psychotherapists. \\
\textbf{Original explanations}: the scientists who affected the colorists are still the scientists. \\
\textbf{BERT explanations}: the scientists are the subject of the relative clause who addressed the psychotherapists and the scientists helped the colorists, we do not know whether the scientists helped the psychotherapists. \\
\midrule
\misvmist \\
\textbf{Premise}: the professor who deceived the athletes called the doctors. \\
\textbf{Hypothesis}: the professor called the doctors. \\
\textbf{Original explanations}: the professor who deceived the athletes is still the professor. \\
\textbf{BERT explanations}: the psychiatrists who the president offended is still the psychiatrist. \\
\bottomrule
\end{tabular}
\caption{Example explanations for all test case generated by the BERT model trained with $k=4$.}
\label{tab:appendix_generated_expl}
\end{table}
\newpage

\begin{table}[ht]
\small
\centering
\begin{tabular}{p{0.45\textwidth}}
\toprule
\mvmt \\
\textbf{Premise}: the directors in front of the analysts avoided the designer. \\
\textbf{Hypothesis}: the directors avoided the designer. \\
\textbf{Original explanations}: the directors in front of the analysts are still the directors. \\
\textbf{BERT explanations}: the directors in front of the analysts are still the directors. \\
\midrule
\misvmt \\
\textbf{Premise}: the chaplains near the singer needed the author. \\
\textbf{Hypothesis}: the chaplains needed the author. \\
\textbf{Original explanations}: the chaplains near the singer are still the chaplains. \\
\textbf{BERT explanations}: the psychologists are in front of the musician and the strategists helped the writer, we do not know whether the illustrators helped the writer. \\
\midrule
\mvmist \\
\textbf{Premise}: the technician that thanked the planners encouraged the worker. \\
\textbf{Hypothesis}: the technician encouraged the worker. \\
\textbf{Original explanations}: the technician that thanked the planners is still the technician. \\
\textbf{BERT explanations}: the technician that encouraged the worker is still the technician, so the technician encouraged the worker and we do not know whether the technician encouraged the worker. \\
\midrule
\misvmist \\
\textbf{Premise}: the senators that recognized the nurses recommended the chaplains. \\
\textbf{Hypothesis}: the senators recommended the chaplains. \\
\textbf{Original explanations}: the senators that recognized the nurses are still the senators. \\
\textbf{BERT explanations}: the presidents are the subject of the relative clause that offended the psychiatrists and the presidents advised the stylists, we do not know whether the officers advised the bakers. \\
\bottomrule
\end{tabular}
\caption{Example explanations for all test case generated by the BERT model trained with $k=16$.}
\label{tab:appendix_k=16_generated_expl}
\end{table}

\end{document}